\documentclass[11pt]{article}

\usepackage[margin=1in]{geometry}
\usepackage{times}
\usepackage{amsmath,amssymb,amsthm}
\usepackage{booktabs}
\usepackage{graphicx}
\usepackage{subcaption}
\usepackage{hyperref}
\usepackage{xcolor}
\usepackage{microtype}
\usepackage{setspace}
\usepackage[numbers,sort&compress]{natbib}

\hypersetup{
  colorlinks=true,
  linkcolor=black,
  citecolor=blue!50!black,
  urlcolor=blue!50!black
}

\title{\textbf{Unthrottling the Tanh Jacobian in SAC:\\A Negative Result on Bang-Bang Control and MetaDrive}}
\author{Faiq Shamass, San Diego, CA, USA\\[0.35em]
{\small Independent Researcher
(e-mail: faiq.shamass@gmail.com).}}
\date{}

\begin{document}
\maketitle

\begin{abstract}
Soft Actor-Critic (SAC) represents a continuous policy as an unbounded Gaussian that is squashed by $\tanh$. The Jacobian of that map is $\partial a/\partial u = 1-a^2$, which vanishes as $|a|\to 1$. A natural concern is that this throttle starves the actor of critic signal exactly where extreme actions---full brake, full throttle---are optimal. We test a minimal intervention that restores the missing signal: one extra term in the actor loss whose gradient on the pre-tanh mean is the detached action-gradient of $Q$, with no gain parameter.

On a minimum-time double integrator whose optimum is bang-bang at the action bounds, vanilla SAC already reaches near-optimal return ($-31.6$ vs.\ a calibrated optimum of $-30.3$) across ten paired seeds. An \emph{ungated} bypass does saturate the policy ($99\%$ of eval steps with $|a|\ge 0.9$) and collapses return to $-195.5$. A \emph{gated} bypass that fires only on the flat shoulder $|a|\in[0.9,0.999]$ also fails, and does so \emph{without} leaving a saturated policy. Warm-started MetaDrive fine-tuning shows the same pattern: the bypass does not improve return, and where collision rate falls it is typically traded for out-of-road departures. Auto-tuned entropy coefficient rises against the bypass, which is a push toward the tails.

The Jacobian effect is real. Treating it as a bug to be undone is not free, and on the tasks studied here it is not helpful. Saturating a bound is not the same as solving a problem whose optimum lives on that bound.
\end{abstract}

\section{Introduction}

Bounded continuous control almost always routes an unbounded policy through a squash. In SAC~\citep{haarnoja2018sac,haarnoja2018algorithms} that squash is $\tanh$, chosen so that the change-of-variables correction for $\log\pi(a\mid s)$ stays closed form~\citep{haarnoja2018sac}. The same map multiplies every critic gradient that reaches the pre-tanh mean $u$ by $1-a^2$. At $|a|=0.9$ that factor is $0.19$; at $|a|=0.99$ it is $0.02$.

If a task rewards sitting on the bound---emergency braking, bang-bang thrust---this looks like a defect. The critic can prefer a more extreme action while the actor receives almost no update. We isolate that hypothesis and test the smallest repair that follows from it:

\begin{quote}
Add a term to the SAC actor loss whose gradient on $u$ is $\partial Q/\partial a$, \emph{not} $(\partial Q/\partial a)\,(1-a^2)$.
\end{quote}

No step-size knob is introduced. On a dimension that is already near saturation, the extra term is stronger than the throttled SAC term by the geometry $1/(1-a^2)$.

The hypothesis does not hold on the two domains we use to test it.

\begin{enumerate}
  \item A one-dimensional cart (minimum-time double integrator) whose every optimum is at $\pm 1$. Vanilla SAC solves it to within a few percent of a hand-calibrated bang-bang reference. Unthrottling the Jacobian either locks the policy to the wall without switching, or---when gated to the shoulder---destabilizes learning without producing a more saturated policy.
  \item Warm-started SAC on MetaDrive~\citep{li2022metadrive}, continued from a checkpoint that already succeeds at low traffic density. Matched baseline/bypass arms under several entropy treatments do not show a return gain attributable to the bypass. Collision reductions, when they appear, come with higher out-of-road rates.
\end{enumerate}

This is a negative result about a specific intervention, not a claim that $\tanh$ is optimal among bound handlers. Alternative action distributions such as the Beta policy~\citep{chou2017beta} remain open. The narrower conclusion is that \emph{re-injecting} the vanished Jacobian into an otherwise standard SAC actor is a different objective, not a faster path to the same one.

\section{Background}

\subsection{Squashed Gaussian policies}

SAC samples $u\sim\mathcal{N}(\mu_\theta(s),\sigma_\theta(s))$ and sets $a=\tanh(u)$. The density on $(-1,1)$ follows from the change of variables~\citep{haarnoja2018sac},
\begin{equation}
  \log\pi(a\mid s)=\log\mu(u\mid s)-\sum_d\log\bigl(1-\tanh^2(u_d)\bigr).
\end{equation}
The Jacobian $da/du=1-a^2$ therefore appears twice: in the likelihood correction, and in the reparameterized actor gradient
\begin{equation}
  \nabla_u Q(s,a)=\bigl(\nabla_a Q(s,a)\bigr)\,(1-a^2).
\end{equation}
Near the bound the second factor vanishes. Entropy regularization~\citep{haarnoja2018sac} pushes $u$ back from the tails and is one reason a squashed Gaussian remains explorable. The same tension is the one we will see empirically: a term that drives $u$ into the tails is opposed by a rising temperature $\alpha$.

\subsection{What the throttle does---and does not---do}

Multiplying a gradient by a small positive number slows travel along that coordinate. It does not, by itself, move the stationary point of the \emph{standard} SAC actor objective, because the same factor scales every contribution that flows through $a=\tanh(u)$. A policy that has already settled where $\nabla_a Q$ balances the entropy term should not sit at a systematically interior action merely because $1-a^2$ is small.

Adding an \emph{unthrottled} copy of $\nabla_a Q$ is different. It changes the first-order condition. The actor is then trained on a mixture of the squashed gradient and the raw action-gradient, which biases $u$ toward whichever bound $Q$ currently points at. Whether that bias is useful depends on whether ``more wall'' is the error the task is making.

\section{Gradient bypass}

Let $u=\mu_\theta(s)$ be the pre-tanh mean head and $a_\mu=\tanh(u)$. Write $w=\nabla_a Q_{\min}(s,a)\big|_{a=a_\mu}$, detached from the critic graph, and let $m\in\{0,1\}^A$ be an optional per-dimension mask. The actor loss is
\begin{equation}
  \mathcal{L}_{\mathrm{actor}}
  =\mathbb{E}\bigl[\alpha\log\pi(a\mid s)-Q(s,a)\bigr]
  -\mathbb{E}\Bigl[\sum_d m_d\,w_d\,u_d\Bigr].
\label{eq:loss}
\end{equation}
The second expectation is the bypass. Because $w$ and $m$ are detached, its gradient on dimension $d$ is exactly $-m_d w_d$ (up to batch averaging). Combined with the standard term,
\begin{equation}
  \frac{\partial\mathcal{L}}{\partial u_d}
  =-w_d(1-a_d^2)\;-\;m_d w_d
\label{eq:grad}
\end{equation}
on a saturated coordinate. There is no scalar gain $\eta$. The relative strength of the extra term is $1/(1-a^2)$ wherever $m_d=1$.

\paragraph{Two masks.}
\emph{Ungated:} $m\equiv 1$. This is the intervention implied by the geometry.
\emph{Gated:} $m_d=1$ only if the current mean sits in the flat shoulder $a_{\mathrm{on}}\le|a_{\mu,d}|<1-\delta$ (we use $a_{\mathrm{on}}=0.9$, $\delta=10^{-3}$). On MetaDrive the gate is further multiplied by a per-dimension blame bit written on failure windows (collision or out-of-road) when the \emph{mean} action was in that shoulder at collection time. The gated variant is what the driving runs use; the toy reports both.

\paragraph{Entropy modes.}
Default SAC auto-tunes $\alpha$. We also run $\alpha=0$ and $\alpha$ frozen at the warm-checkpoint value, because a push into the tails is an anti-entropy force and the tuner can cancel it.

\paragraph{Implementation.}
Both arms use the same \texttt{SACBypass} class, a thin subclass of Stable-Baselines3 SAC~\citep{raffin2021sb3}. \texttt{bypass\_off=True} drops the second term of~\eqref{eq:loss} and leaves every other choice identical: collection, replay, critic, targets, batch size.

\section{Toy: minimum-time double integrator}

\subsection{Task}

A frictionless cart with state $(x,v)$ and scalar action $a\in[-1,1]$ obeys $v\leftarrow v+a\,\Delta t$, $x\leftarrow x+v\,\Delta t$ with $\Delta t=0.05$. Episodes start from $x\sim\mathrm{Unif}[-1,1]$, $v\sim\mathrm{Unif}[-0.5,0.5]$ and end when $\max(|x|,|v|)<0.1$ or after $200$ steps. The true objective is latency: $-1$ per step. We add a potential-based shaping term $\gamma\Phi(s')-\Phi(s)$ with $\Phi(s)=-2(|x|+|v|)$, which leaves the optimal policy unchanged~\citep{ng1999shaping}. Reported scores are always unshaped step counts.

The optimum is bang-bang: full thrust one way, then a single switch to full thrust the other way~\citep{pontryagin1962}. Intermediate actions are never optimal. If the tanh throttle were a first-order obstacle, this is where it should show.

A family of reference controllers plays the optimal switching law subject to an artificial thrust cap $|a|\le c$. On a fixed 44-start eval grid their returns are
\[
\begin{array}{rccccc}
c & 1.0 & 0.9 & 0.7 & 0.5 & 0.3\\
\hline
\mathrm{return} & -30.3 & -31.5 & -36.2 & -43.7 & -60.0.
\end{array}
\]
A learned score can therefore be read back as an \emph{effective action cap}.

\subsection{Protocol}

Ten paired seeds, $40{,}000$ environment steps, evaluation every $2{,}500$ steps on the same 44 starts. Both arms use the production \texttt{SACBypass} class. The gated arm uses $a_{\mathrm{on}}=0.9$; the ungated arm sets $a_{\mathrm{on}}=0$ and blame $=1$ on every transition. The threshold used for sample-efficiency is return $\ge -35$.

\subsection{Results}

\begin{table}[t]
\centering
\caption{Toy double integrator, $10$ paired seeds, $40$k steps. Means over seeds. Paired $t$ is descriptive: seeds are independent, eval starts are shared.}
\label{tab:toy}
\resizebox{\textwidth}{!}{%
\begin{tabular}{lrrrrrr}
\toprule
 & \multicolumn{3}{c}{Gated} & \multicolumn{3}{c}{Ungated} \\
\cmidrule(lr){2-4}\cmidrule(lr){5-7}
 & baseline & bypass & $\Delta$ & baseline & bypass & $\Delta$ \\
\midrule
Final return & $-31.6$ & $-108.6$ & $-76.9$ & $-31.6$ & $-195.5$ & $-163.9$ \\
Reach rate & $1.00$ & $0.58$ & $-0.42$ & $1.00$ & $0.02$ & $-0.98$ \\
Steps to $-35$ & $22{,}500$ & $>40{,}000$ & $+17{,}500$ & $22{,}500$ & $>40{,}000$ & $+17{,}500$ \\
Mean $|a|$ & $0.69$ & $0.17$ & $-0.53$ & $0.69$ & $1.00$ & $+0.30$ \\
Sat.\ frac.\ $|a|\ge 0.9$ & $0.32$ & $0.05$ & $-0.27$ & $0.32$ & $0.99$ & $+0.67$ \\
$p_{95}(|a|)$ & $0.97$ & $0.70$ & $-0.27$ & $0.97$ & $1.00$ & $+0.03$ \\
Effective cap & $\approx 0.89$ & $\approx 0.30$ &  & $\approx 0.89$ & $\approx 0.30$ &  \\
\bottomrule
\end{tabular}}
\end{table}

\begin{figure}[t]
\centering
\includegraphics[width=\textwidth]{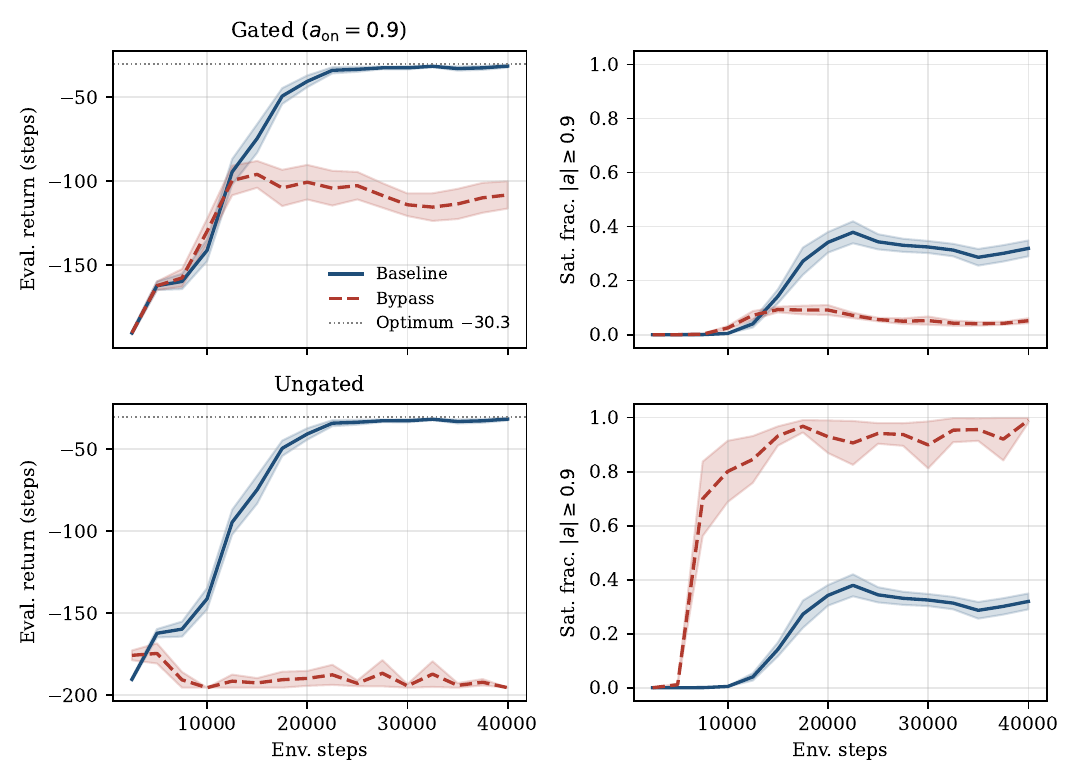}
\caption{Toy learning curves (mean $\pm$ s.e.\ over $10$ seeds). Left: unshaped eval return. Dotted line is the uncapped bang-bang reference ($-30.3$). Right: fraction of eval steps with $|a|\ge 0.9$. Top: gated bypass. Bottom: ungated bypass. Vanilla SAC reaches the reference band; both bypasses do not.}
\label{fig:toy}
\end{figure}

Vanilla SAC is already close to the bound. Final return $-31.6$ matches the $c=0.9$ reference; $p_{95}(|a|)=0.97$. The original worry---that the policy would stall near $|a|=0.9$ and stay there---does not describe this baseline.

The two bypasses fail for opposite mechanistic reasons (Table~\ref{tab:toy}, Figure~\ref{fig:toy}).

\paragraph{Ungated.}
The term does what the algebra says. Saturation fraction goes from $0.32$ to $0.99$ and mean $|a|$ goes to $1.00$. Return collapses to $-195.5$ and reach rate to $2\%$. The policy is on the wall and does not solve the switching problem. Reading the score through the calibration table gives an effective cap of $\approx 0.3$, not because the action is small, but because a policy that is always saturated in the wrong direction is as slow as a weak actuator.

\paragraph{Gated.}
The shoulder gate prevents the lock-to-wall seen above. It also prevents the intended mechanism from appearing at evaluation: saturation \emph{falls} relative to baseline ($0.32\to 0.05$). Return is still far worse ($-108.6$, reach $58\%$). The gate therefore does not convert a harmful term into a harmless one. It produces a third outcome---an unsaturated, low-quality policy---rather than recovering vanilla SAC.

In both cases the sample-efficiency hypothesis (same destination, shorter path) is rejected: no bypass seed reaches the $-35$ threshold inside $40$k steps; every baseline seed does.

\section{MetaDrive fine-tuning}

\subsection{Setup}

We warm-start from a vanilla-SAC checkpoint trained on MetaDrive at traffic density $0.1$, where that checkpoint is already strong (held-out success $\approx 0.92$, return $\approx 301$ at density $0.1$). Fine-tuning uses the same codebase with a manual collect-and-train loop, one gradient step per environment step, and deterministic held-out evaluation on a fixed seed block. Most runs use density $0.3$ so that the warm policy still fails often enough for a gain to be visible; a frozen-$\alpha$ pair is also repeated at density $0.1$.

The driving bypass is the \emph{gated} form, with blame written on the last $100$ transitions of a failed episode only when the mean action was in the shoulder. Entropy is left auto-tuned, turned off ($\alpha=0$), or frozen at the checkpoint value $\alpha\approx 0.023$. Eval uses $20$ episodes on the early entropy-on/off pair and $50$ episodes on the frozen-$\alpha$ pair. Unless noted, driving comparisons are a single paired seed.

\subsection{How we read the curves}

Consecutive evals are correlated, so we do not treat $101$ eval points as $101$ independent trials. We report two complementary summaries: the mean of the last ten evals (``where the arm ended''), and the mean paired difference $\Delta_k = y_k - b_k$ after learning has started (``did the term help across training''). The paired $t$ on those differences is a ranking signal, not a $p$-value.

\begin{table}[t]
\centering
\caption{MetaDrive warm-start. Last-ten eval means, then paired $\Delta$ (bypass $-$ baseline) over post-warmup evals. Frozen-$\alpha$ pairs have $101$ evals; entropy on/off pairs have $21$ evals at $20$ episodes each. Extra $\alpha=0$ bypass seeds $50$ and $100$ have no matched baseline.}
\label{tab:md}
\small
\begin{tabular}{llrrrr}
\toprule
Setting & arm & return & success & collision & out-of-road \\
\midrule
Entropy ON, td $0.3$
  & base & $161.6$ & $0.19$ & $0.39$ & $0.10$ \\
  & bypass & $154.7$ & $0.16$ & $0.31$ & $0.17$ \\
  & paired $\Delta$ & $-3.9$ & $-0.02$ & $-0.08$ & $+0.07$ \\
\midrule
Entropy OFF, td $0.3$
  & base & $138.1$ & $0.09$ & $0.31$ & $0.24$ \\
  & bypass & $151.2$ & $0.13$ & $0.20$ & $0.27$ \\
  & paired $\Delta$ & $+10.8$ & $+0.03$ & $-0.07$ & $+0.00$ \\
\midrule
$\alpha$ frozen, td $0.3$
  & base & $135.4$ & $0.11$ & $0.39$ & $0.24$ \\
  & bypass & $130.7$ & $0.09$ & $0.37$ & $0.28$ \\
  & paired $\Delta$ & $-3.2$ & $-0.01$ & $-0.02$ & $+0.03$ \\
\midrule
$\alpha$ frozen, td $0.1$
  & base & $265.5$ & $0.77$ & $0.09$ & $0.14$ \\
  & bypass & $240.4$ & $0.70$ & $0.05$ & $0.24$ \\
  & paired $\Delta$ & $-2.4$ & $+0.02$ & $-0.02$ & $+0.00$ \\
\bottomrule
\end{tabular}
\end{table}

\begin{figure}[t]
\centering
\includegraphics[width=\textwidth]{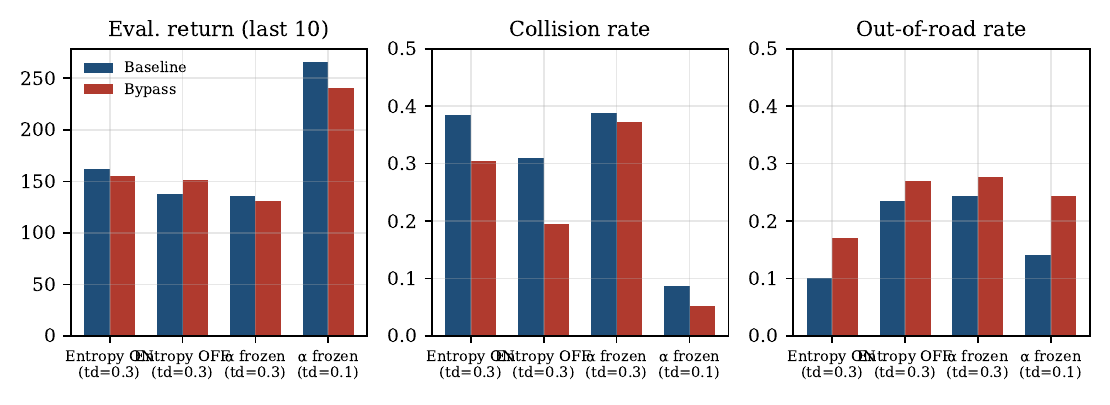}
\caption{MetaDrive last-ten-eval snapshot for the four matched pairs. The bypass never produces a clean return win. Collision drops, when present, track higher out-of-road rates in three of four settings.}
\label{fig:md}
\end{figure}

\subsection{Results}

All density-$0.3$ arms \emph{decay} from the warm checkpoint (success $0.28$ or $0.10$ at $t{=}0$ depending on eval protocol, then lower). The comparison is therefore about rates of decay, not about lifting a solved task.

\paragraph{Entropy on.}
Bypass is slightly worse on return and success. Collision rate falls; out-of-road rises by a similar amount. That is a mode trade, not an improvement in driving.

\paragraph{Entropy off.}
On seed $0$ the bypass beats its no-entropy baseline ($+10.8$ return). Two caveats block a method claim. First, $\alpha=0$ also hurts the \emph{baseline} relative to entropy-on SAC ($138$ vs.\ $162$), so the apparent bypass gain is against a weakened control, not against standard SAC. Second, two further no-entropy bypass seeds land in different basins (last-ten return $132$ and $159$; out-of-road $0.36$ and $0.08$) and have no seed-matched baseline.

\paragraph{Frozen $\alpha$.}
At density $0.3$ the paired return delta is negative. At density $0.1$ return is tied across training and the last window trades collisions ($0.086\to 0.052$) for out-of-road ($0.140\to 0.244$), while both arms sit well below the $t{=}0$ checkpoint (success $0.92\to\sim 0.70$--$0.77$).

Taken together, MetaDrive does not rescue the toy. The only cell that looks locally positive is $\alpha=0$ on one seed, and that cell is the least standard SAC.

\section{Discussion}

Three observations survive the experiments.

\paragraph{The baseline was not Jacobian-starved.}
On the task that most favors the hypothesis, vanilla SAC already commands $p_{95}(|a|)\approx 0.97$ and matches a $0.9$-thrust reference. ``The policy cannot reach the bound'' is the wrong diagnosis of its remaining error.

\paragraph{Unthrottling changes the objective.}
Equation~\eqref{eq:grad} is not a rescaled SAC update. The extra $-w$ term keeps paying for motion toward the bound after $1-a^2$ has gone to zero. On a bang-bang problem the hard part is the switch, not the first contact with $\pm 1$. A force that rewards remaining in the tail fights that switch. The ungated toy is the clean picture of this: saturation up, return down.

\paragraph{Entropy is not a spectator.}
A mean driven into the tails is a more deterministic actor. Auto-tuned $\alpha$ rises to compensate, which is why the entropy-on driving pair is flat-to-negative and why $\alpha=0$ is the only cell with a local bypass edge. Removing entropy to ``let the bypass work'' discards a stabilizer that the baseline needs.

A more plausible research direction than another gate is a different bound representation---Beta policies~\citep{chou2017beta}, or a residual around a saturated controller that already knows when to switch. Those change how bounds are represented. They do not re-inject a vanished $\partial a/\partial u$ into an unchanged squash.

\section{Limitations}

Driving results are mostly one paired seed; the extra no-entropy seeds lack matched baselines. Fine-tuning starts from a strong checkpoint and, at density $0.3$, from a shifted traffic distribution; both arms often finish below $t{=}0$. Early driving evals use $20$ episodes. The MetaDrive bypass is blame-and-band gated, so it is not the same object as the ungated toy term. We did not run from-scratch MetaDrive, throttle-only ablations, or a multi-seed frozen-$\alpha$ grid to completion. None of these caveats revive the toy result, which is the stronger and cleaner negative.

\section{Conclusion}

We asked whether restoring the critic gradient that $\tanh$ attenuates near $\pm 1$ would help SAC on tasks that want extreme actions. On a minimum-time double integrator it does not: vanilla SAC is already near the bound, an ungated bypass saturates the action and destroys switching, and a gated bypass fails without even saturating. On MetaDrive fine-tuning the same term does not improve return.

The Jacobian of $\tanh$ is a real factor in the actor gradient. It is not, on this evidence, the reason these policies fail, and undoing it is not a free speedup. Code and logs are available at \url{https://github.com/fshamass/Tanh-Bypass}.

\bibliographystyle{plainnat}

\begin{thebibliography}{99}

\bibitem[Chou et~al.(2017)]{chou2017beta}
P.-W. Chou, D. Maturana, and S. Scherer.
\newblock Improving stochastic policy gradients in continuous control with the Beta distribution.
\newblock In \emph{ICLR}, 2017.

\bibitem[Haarnoja et~al.(2018a)]{haarnoja2018sac}
T. Haarnoja, A. Zhou, P. Abbeel, and S. Levine.
\newblock Soft actor-critic: Off-policy maximum entropy deep reinforcement learning with a stochastic actor.
\newblock In \emph{ICML}, 2018.

\bibitem[Haarnoja et~al.(2018b)]{haarnoja2018algorithms}
T. Haarnoja, A. Zhou, K. Hartikainen, G. Tucker, S. Ha, J. Tan, V. Kumar, H. Zhu, A. Gupta, P. Abbeel, and S. Levine.
\newblock Soft actor-critic algorithms and applications.
\newblock \emph{arXiv:1812.05905}, 2018.

\bibitem[Li et~al.(2022)]{li2022metadrive}
Q. Li, Z. Peng, L. Feng, Q. Zhang, Z. Xue, and B. Zhou.
\newblock MetaDrive: Composing diverse driving scenarios for generalizable reinforcement learning.
\newblock \emph{IEEE TPAMI} / CoRL, 2022.

\bibitem[Ng et~al.(1999)]{ng1999shaping}
A.~Y. Ng, D. Harada, and S. Russell.
\newblock Policy invariance under reward transformations: Theory and application to reward shaping.
\newblock In \emph{ICML}, 1999.

\bibitem[Pontryagin et~al.(1962)]{pontryagin1962}
L.~S. Pontryagin, V.~G. Boltyanskii, R.~V. Gamkrelidze, and E.~F. Mishchenko.
\newblock \emph{The Mathematical Theory of Optimal Processes}.
\newblock 1962.

\bibitem[Raffin et~al.(2021)]{raffin2021sb3}
A. Raffin, A. Hill, A. Gleave, A. Kanervisto, M. Ernestus, and N. Dormann.
\newblock Stable-Baselines3: Reliable reinforcement learning implementations.
\newblock \emph{Journal of Machine Learning Research}, 22(268):1--8, 2021.

\end{thebibliography}

\end{document}